\documentclass[a4paper,fleqn]{cas-sc}
\usepackage{times}
\usepackage{lineno}

\usepackage[numbers]{natbib}
\usepackage{graphicx}
\usepackage{booktabs}
\usepackage{multirow}
\usepackage[dvipsnames]{xcolor}
\usepackage{color,mdwlist}
\definecolor{c1}{HTML}{b71a3b}
\definecolor{c2}{HTML}{17A1DE}
\definecolor{c3}{HTML}{E85642}
\definecolor{c4}{HTML}{801dae}

\usepackage{subfig}
\usepackage{subcaption}
\usepackage{caption}
\def\tsc#1{\csdef{#1}{\textsc{\lowercase{#1}}\xspace}}
\tsc{WGM}
\tsc{QE}

\begin{document}

\let\WriteBookmarks\relax
\def\floatpagepagefraction{1}
\def\textpagefraction{.001}
\let\printorcid\relax 

\shorttitle{GSO-YOLO: Global Stability Optimization YOLO for Construction Site Detection}    

\shortauthors{Yuming Zhang, Dongzhi Guan, Shouxin Zhang et al.}

\title[mode = title]{GSO-YOLO: Global Stability Optimization YOLO for Construction Site Detection}

\author{Yuming Zhang} 
\author{Dongzhi Guan}
\cormark[1] 
\author{Shouxin Zhang}
\author{Junhao Su}
\author{Yunzhi Han}
\author{Jiabin Liu}

\address{School of Civil Engineering, Southeast University, Nanjing 211189, China}


\begin{abstract}
Safety issues at construction sites have long plagued the industry, posing risks to worker safety and causing economic damage due to potential hazards. With the advancement of artificial intelligence, particularly in the field of computer vision, the automation of safety monitoring on construction sites has emerged as a solution to this longstanding issue. Despite achieving impressive performance, advanced object detection methods like YOLOv8 still face challenges in handling the complex conditions found at construction sites. To solve these problems, this study presents the Global Stability Optimization YOLO (GSO-YOLO) model to address challenges in complex construction sites. The model integrates the Global Optimization Module (GOM) and Steady Capture Module (SCM) to enhance global contextual information capture and detection stability. The innovative AIoU loss function, which combines CIoU and EIoU, improves detection accuracy and efficiency. Experiments on datasets like SODA, MOCS, and CIS show that GSO-YOLO outperforms existing methods, achieving SOTA performance.
\end{abstract}

\begin{keywords}
Object Detection \sep 
Deep Learning \sep 
Construction Site Monitoring \sep
YOLOv8
\end{keywords}
\maketitle

\section{Introduction}
Safety accidents at construction sites not only pose severe threats to the lives of construction workers but also result in significant economic losses \cite{r1}. According to statistics, from 2015 to 2019, the residential and municipal engineering industry in China experienced 3,275 construction safety accidents, resulting in a total of 3,840 deaths, with the number of accidents and fatalities increasing at an average annual rate of 14.98\% and 12.64\%, respectively \cite{r2}. This indicates an increasingly severe trend in construction site safety accidents. Therefore, effectively enhancing safety management and engineering supervision at construction sites to reduce casualties and improve construction efficiency has become a critical issue that the construction industry urgently needs to address \cite{r3}.

Research indicates that more than half of construction safety accidents can be timely prevented through enhanced monitoring of construction sites \cite{r4}. However, the commonly used monitoring methods are primarily traditional techniques such as work sampling and personnel testing \cite{r5}, which not only consume substantial human, material, and financial resources but also suffer from inefficiency and the inability to provide comprehensive, around-the-clock monitoring \cite{r6}. With the development of information and communication technology, some scholars have begun to explore the use of computer vision technology to achieve automated monitoring of construction sites \cite{r7,r8,r9}. For example, Fang et al. \cite{r10} develop an integrated hybrid learning method based on Faster R-CNN and Deep CNN to detect the use of safety harnesses by high-altitude construction workers. Wang et al. \cite{r11} integrate computer vision technology with multispectral data for the detection of construction machinery. However, the current computer vision-based automated monitoring technologies for construction sites still face the following three urgent problems:
\begin{figure}[!t]
\centering
\includegraphics[width=3.15in]{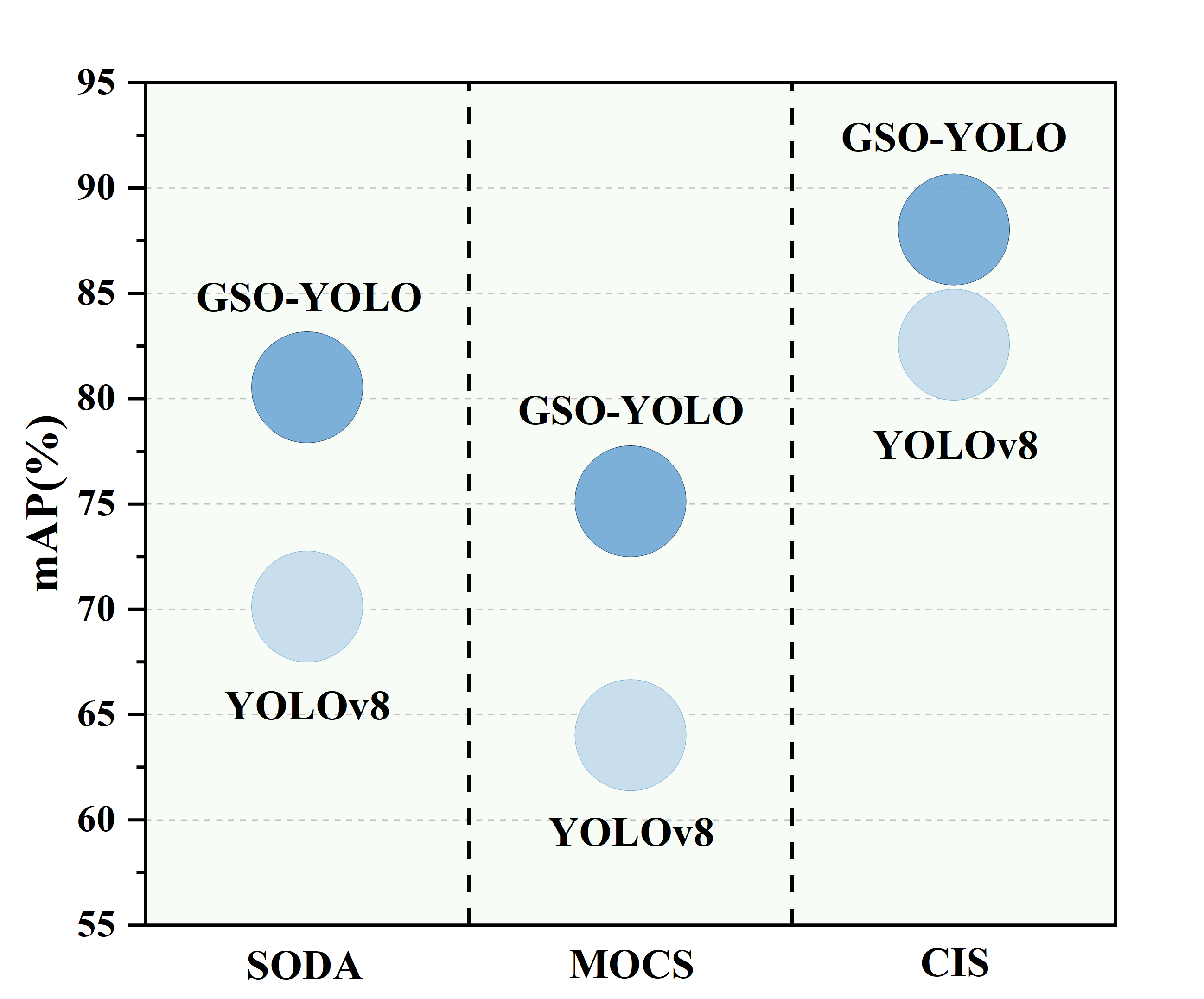}%
\caption{Comparison between GSO-YOLO and YOLOv8.}
\label{fig:1}
\end{figure}

\begin{itemize}
\item Due to the complex backgrounds of construction sites, detection targets are frequently momentarily obstructed by obstacles such as construction vehicles. Consequently, the pixels captured by detection equipment often lack complete features. This results in a significant amount of noise, which severely interferes with the model's ability to extract target features, thereby greatly reducing the accuracy of the detection model.
\item Since the detection targets are exposed to outdoor conditions, variations in lighting, weather, and resolution occur at different times. In low-light environments, the detection model struggles to identify effective feature information about the target objects, thereby reducing the reliability of the model's recognition capability.
\item Given the large size of construction sites and the fact that detection equipment is often installed at wide angles far from the target areas, the detection model finds it challenging to extract feature information from distant targets. This increases the difficulty of accurate detection.
\end{itemize}

To address the aforementioned issues, this study proposes the Global Stability Optimization-YOLO (GSO-YOLO) model, as illustrated in Figure \ref{fig:2}, which integrates the Global Optimization Module (GOM) and the Steady Capture Module (SCM) into the YOLOv8 model to handle the complexities of construction site environments. Additionally, this study has innovatively designed the AIoU loss function. Specifically, the GOM maximizes the capture of relevant information about detection targets within images by applying global attention weighting across input sequences and combining it with local information, thereby enhancing model performance and generalization capabilities. The SCM, on the other hand, smooths detection results and reduces noise interference by dynamically applying an exponentially weighted moving average to historical detection results, thereby improving the model's stability and robustness. The combination of these two modules increases the receptive field of the network and enhances detection accuracy. Furthermore, by combining the advantages of the CIoU \cite{r44} and EIoU \cite{r45} loss functions, this study introduces a novel enhanced loss function, AIoU. With the deployment of GOM, SCM, and AIoU, GSO-YOLO effectively mitigates the interference of complex environments, momentary occlusions, and lighting variations in construction site object detection. Extensive experiments conducted on datasets such as SODA \cite{r46}, MOCS \cite{r47}, and CIS \cite{r48} validate the effectiveness of the proposed method. Overall, contributions are as follows:

\begin{figure*}[!t]
\centering
\includegraphics[width=6.5in]{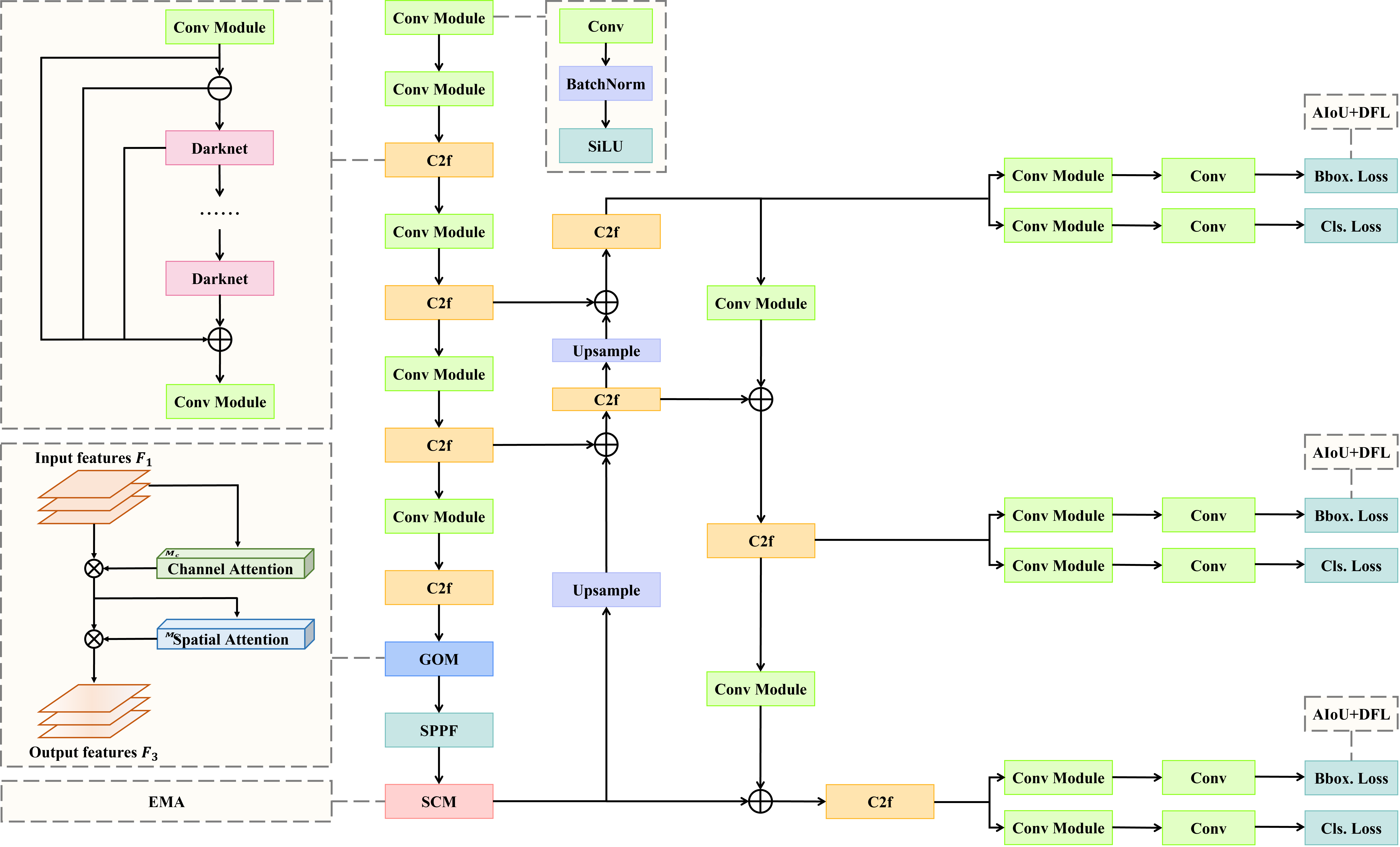}%
\caption{The GSO-YOLO overall architecture.}
\label{fig:2}
\end{figure*}

\begin{itemize}
\item This study presents a novel method called GSO-YOLO, which successfully addresses the shortcomings of existing YOLO series algorithms and effectively mitigates the interference caused by complex environments in construction site object detection. GSO-YOLO enables the network to focus on and learn critical information from the entire construction site globally, increasing the receptive field without being confined to local details. It also enhances the capture of historical information, providing stable tracking and recognition of targets.
\item The Global Optimization Module (GOM) and Steady Capture Module (SCM) designed in this study are general methods with excellent generalizability, making them applicable to various deep learning-based object detection frameworks.
\item GSO-YOLO outperforms existing mainstream methods on multiple public datasets, achieving state-of-the-art (SOTA) performance. This demonstrates its significant potential and reliability in practical applications, substantially advancing the technology of construction site object detection.
\end{itemize}

\section{Related Works}
Currently, leveraging computer vision technology to achieve automated monitoring of construction sites has garnered widespread attention in academia due to its potential to enhance project safety and productivity \cite{r12,r13,r14}. Researchers have primarily focused on various aspects of construction site automation, including progress monitoring \cite{r15,r16}, quality management \cite{r17,r18,r19}, activity recognition \cite{r20,r21}, occupational health assessment \cite{r22,r23}, worker safety inspections \cite{r24,r25,r26}, automatic fire detection \cite{r27}, collision risk prevention \cite{r28}, equipment activity analysis \cite{r29}, and smart glasses systems \cite{r30}. However, most of these studies rely on handcrafted features and sliding window approaches to traverse entire images, resulting in slow algorithm performance, insufficient accuracy, and a lack of model generalizability \cite{r31}.

In contrast, deep learning convolutional neural networks (CNNs) have demonstrated superior performance in object detection tasks \cite{r32,r33}. CNN-based detectors utilize adaptive convolution for feature extraction, which enhances both the accuracy and speed of algorithms while making models more robust in object detection \cite{r4}. For example, some studies have applied CNNs for structural defect detection \cite{r34,r35}, building energy consumption prediction \cite{r36}, and scaffold deformation monitoring \cite{r37}. 

Although progress has been made, these studies face several challenges that limit their effectiveness in complex construction site environments. One significant issue is their reliance on low-level features to improve the detection of small objects, often at the expense of accurately detecting larger objects \cite{r38}. Extending feature levels using low-level features can enhance small object detection accuracy but can simultaneously reduce the accuracy for large object detection \cite{r39,r40}. 

In this regard, the YOLO series algorithms have demonstrated excellent performance. Notably, Hao et al. \cite{r5} propose a lightweight multi-object detection model based on YOLOv5s, achieving notable success in construction site monitoring. Similarly, Han et al. \cite{r38} develop the ghost-YOLOX model for nighttime construction site monitoring.

However, existing YOLO algorithms struggle with noise interference from momentary obstructions and lighting variations at construction sites, which diminishes their performance in these dynamic environments. Despite numerous improvements to the loss functions used in YOLO algorithms, there remains room for further enhancements to address these challenges comprehensively. These limitations underscore the necessity for more robust and versatile approaches in automated construction site monitoring. Thus, there is a need for YOLO series algorithms that balance detection accuracy across small, medium, and large objects.

\section{Method}
The stacking of deep convolutional layers consumes a significant amount of memory and computational resources, prompting the introduction of new modules incorporating attention mechanisms as an alternative approach. These modules not only enhance the learning of more distinctive features but can also be easily integrated into the backbone architecture of neural networks. Based on this, to better adapt the proposed network model to the requirements of construction site object detection, this study has incorporated attention mechanism modules into the backbone network of YOLOv8. This integration aims to improve the accuracy of object detection without merely increasing the network depth by adding more convolutional layers. 

\subsection{Global Optimization Module}

In the Global Optimization Module (GOM), this study utilizes the Global Attention Mechanism (GAM)\cite{r42} to apply attention weighting across the global input sequence. This mechanism extracts global contextual information and integrates it with local information for computation. GAM simultaneously considers three-dimensional attention weights—channel, spatial width, and spatial height. Channel attention employs a 3D arrangement to maintain information across dimensions, using two layers of multi-layer perceptrons (MLP) to amplify spatial correlations across channels. To focus on spatial information, the GAM uses two convolutional layers for spatial information fusion and eliminates the pooling process to preserve feature maps.

Assuming the input feature map is \( F_1 \in \mathbb{R}^{C \times H \times W} \), the intermediate state is \( F_2 \), and the output feature map is \( F_3 \). They can be represented as follows:

\nolinenumbers
\begin{equation}
    F_{2}=M_{c}\left(F_{1}\right) \otimes F_{1}
\end{equation}
\begin{equation}
    F_{3}=M_{s}\left(F_{2}\right) \otimes F_{2}
\end{equation}
Assuming \( M_{c} \) and \( M_{s} \) represent the channel and spatial attention maps, respectively, and \( \otimes \) denotes element-wise multiplication.

The attention parameters of the GAM are updated through gradient descent, represented as follows:
\begin{equation}
    \theta_{i}=\theta_{i-1}-r \times \nabla_{\theta} L
\end{equation}
Here, \(\theta_i\) is the attention weight parameter after the \(i\)-th iteration, \(\theta_{(i-1)}\) is the attention weight parameter from the previous iteration, \(r\) is the learning rate, \(L\) is the loss function, and \(\nabla_{\theta} L\) is the gradient of the loss function concerning the attention weight parameter, which can be obtained through backpropagation.

This study's objective is to enable the entire network model to mitigate information loss and enhance global dimension interaction features as much as possible with the assistance of the Global Optimization Module (GOM). By considering spatial interactions and supplementing cross-dimensional information, the model can focus on the most relevant parts of the sequence, thereby improving performance and generalization capabilities. In construction sites, there are typically various complex scenes and objects, such as building structures, workers, and machinery. To effectively identify and understand these objects, GOM can assist the object detection network in global optimization. Specifically, GOM enables the network to globally focus on and learn the key information of the entire construction site, expanding the receptive field without being limited to local details. This allows for a more comprehensive understanding and identification of various objects at the construction site. This capability to integrate semantic information and structural features enables the object detection model to more accurately reflect the real conditions of the construction site, providing robust technical support for construction management and safety monitoring.

\subsection{Steady Capture Module}

In the Steady Capture Module (SCM), this study employs the Exponential Moving Average (EMA) \cite{r43}. EMA is an attention mechanism used to enhance the performance of deep learning models by dynamically adjusting weights to emphasize the influence of recent data, thereby adapting to changes in the input data. Specifically, each input feature is assigned a weight that determines its importance in the output computation. These weights are typically generated by a softmax function, with the input being a learned score vector.

During training, as weights are passed into the Exponential Moving Average (EMA), these score vectors are dynamically adjusted. A neural network model with EMA will dynamically update the attention weights based on the feedback from the loss function, allowing it to better capture the relevance of the input data. Moreover, the EMA attention mechanism introduces a decay rate to control the degree of forgetfulness regarding historical information. The smaller the decay rate, the higher the model's dependency on historical information, and vice versa. This decay rate typically ranges between 0 and 1.

In EMA, the formula for calculating attention weights can be expressed as follows:
\begin{equation}
    w_{i}(t)=\frac{e^{s_{i}(t)}}{\sum_{j=1}^{N} e^{s_{j}(t)}}
\end{equation}
where \( w_i(t) \) represents the attention weight of the \( i \)-th input, and \( s_j(t) \) is the score of the \( i \)-th input calculated by the neural network. During training, \( s_j(t) \) is dynamically adjusted. The update rule is expressed as:
\begin{equation}
    s_{j}(t)=(1-\alpha) \times s_{i}(t-1)+\alpha \times s_{i}^{\prime}(t)
\end{equation}
where \(\alpha\) represents the decay rate, which controls the retention of historical information, and \(s_i'(t)\) is the current score at time \(t\).

In the task of object detection at construction sites, the complexity and dynamism of the environment can lead to challenges such as temporary occlusions and varying lighting conditions, making it difficult for detection models to accurately identify objects. To address this, this study employs the SCM to help the model better capture historical information and achieves stable tracking and recognition of targets.

In the SCM, the use of the EMA significantly impacts both the model's performance and its stability during training. On one hand, EMA controls the retention of historical information through the decay rate \(\alpha\). Historical information refers to the feature data from previously processed images or image sequences. Construction sites may contain various background elements, such as buildings and machinery, which, although not targets, assist the model in understanding the positions and shapes of current targets. However, excessive reliance on historical information might lead to an inaccurate understanding of the current scene. The decay rate \(\alpha\) serves as a balancing factor, controlling the model’s dependency on historical information, allowing it to retain past data to some extent while also adapting to new scenes in a timely manner.

On the other hand, construction site datasets often contain noise and uncertainties, such as image blurring or distortion caused by weather and lighting variations. In such cases, the model needs to maintain robustness, performing well even in complex environments. In the SCM, adjusting \(\alpha\) helps the model better handle these uncertainties, thereby enhancing its robustness. By fine-tuning \(\alpha\), the model achieves an optimal balance between historical data retention and adaptability to new information, ensuring consistent and accurate object detection in construction site environments. 

\subsection{Augmented Intersection over Union}

In this research, a novel augmented loss function called AIoU is used, which integrates the CIoU \cite{r44} and EIoU \cite{r45} loss functions as its foundation. In YOLOv8, the regression loss function uses CIoU \cite{r44}, and the CIoU loss function is expressed as follows:
\begin{equation}
    \mathcal L_{\text {CIoU }}=1-I o U+\frac{\rho^{2}\left(b, b^{g t}\right)}{c^{2}}+\alpha v
\end{equation}
In this context, \( b \) and \( b^{gt} \) represent the center points of the predicted box and the target box, respectively. \( \rho(\cdot) \) denotes the Euclidean distance, \( c \) is the diagonal length of the smallest enclosing box that covers both boxes, \( \alpha \) is a positive trade-off parameter, and \( \nu \) is the consistency of the aspect ratio.

In the formula (6), \( \nu \) reflects the difference in aspect ratios rather than the actual differences in width and height with their respective confidences. Consequently, it sometimes hinders the model's effective optimization of similarity. The penalty term in EIoU disaggregates the influence factors of the aspect ratio, calculating the length and width of the target box and the predicted box separately based on the penalty term in CIoU. The loss function includes three parts: overlap loss, center distance loss, and width-height loss. The first two parts follow the methods in CIoU, but the width-height loss directly minimizes the difference in width and height between the target box and the predicted box, resulting in faster convergence. The EIoU loss function is expressed as follows:
\begin{equation}
    \mathcal L_{E I o U}=1-I o U+\frac{\rho^{2}\left(b, b^{g t}\right)}{c_{w}{ }^{2}+c_{h}{ }^{2}}+\frac{\rho^{2}\left(w, w^{g t}\right)}{c_{w}{ }^{2}}+\frac{\rho^{2}\left(b, b^{g t}\right)}{c_{h}{ }^{2}}
\end{equation}

In the application of AIoU, the aspect ratio of the bounding box is first adjusted by CIoU until it converges to an appropriate range. Then, each edge is meticulously refined by EIoU until it converges to the correct value. The AIoU loss is computed using the following formula:
\begin{equation}
    \mathcal L_{A I o U}=1-I o U+\alpha v+\frac{\rho^{2}\left(b, b^{g t}\right)}{c_{w}^{2}+c_{h}^{2}}+\frac{\rho^{2}\left(w, w^{g t}\right)}{c_{w}^{2}}+\frac{\rho^{2}\left(h, h^{g t}\right)}{c_{h}^{2}}
\end{equation}

CIoU emphasizes the completeness of the intersection and union between target detection boxes, while EIoU focuses on computational efficiency. Combining the two allows for a comprehensive consideration of both detection accuracy and computational efficiency, enabling the model to maintain high accuracy while utilizing computational resources more effectively. In construction site environments, precise object location information is crucial for monitoring and safety management, and computational efficiency is also an important factor. The combination of CIoU and EIoU balances these needs: CIoU enhances sensitivity in detecting small objects, and EIoU increases the model's robustness, allowing it to better adapt to complex environmental changes.

\subsection{Overall Architecture Adjustment}

This study uses YOLOv8 as the primary network for the experiments and has made several key improvements and adjustments to its structure. First, it introduces the GOM module before the SPPF module. This design decision is based on GOM's ability to enhance the network's perceptual field, allowing it to better understand the context and information of the entire image, thereby improving object detection accuracy and robustness. Second, it proposes the SCM module immediately after the SPPF. The role of this module is to further enhance object localization and capture based on the multi-scale features extracted by the SPPF, particularly for detecting small or low-contrast objects. Finally, it utilizes AIoU throughout the main network to propagate loss. AIoU is an adaptive IoU computation method that more accurately assesses detection accuracy and helps the network learn the precise boundaries of objects during training. This design effectively improves the model's performance on construction site datasets, making it more suitable for real-world construction scenarios by enhancing the accuracy and robustness of object detection.

\section{Experiment} 
\subsection{Datasets}

In this section, this study briefly introduces the three large-scale construction site datasets it uses: SODA \cite{r46}, MOCS \cite{r47}, and CIS \cite{r48}.

The SODA dataset is an image dataset in VOC format, containing 19,846 images and annotations for 286,201 objects. SODA includes 15 categories: slogan, fence, hook, hopper, electric box, cutter, handcart, scaffold, brick, rebar, wood, board, helmet, vest, and person, covering the most common objects found on construction sites.

The MOCS dataset is a large-scale image dataset for detecting moving objects in construction sites. All images in the MOCS dataset are collected from real construction sites. The dataset comprises 41,668 images with annotations for 13 categories of objects. The categories in MOCS are worker, tower crane, hanging hook, vehicle crane, roller, bulldozer, excavator, truck, loader, pump truck, concrete transport mixer, pile driver, and other vehicles.

The CIS dataset is a novel image dataset designed to advance state-of-the-art instance segmentation in the field of construction management. It contains 50,000 images with 10 object categories, including two categories of workers: workers wearing and not wearing safety helmets; one category of materials: precast components (PCs); and seven categories of machines: PC delivery trucks, dump trucks, concrete mixer trucks, excavators, rollers, dozers, and wheel loaders.

\subsection{Implement Detail}

In the experiments, this study uses an RTX-3060 GPU as the experimental platform and selects YOLOv8 as the primary backbone network. It conducts experiments on three widely used construction site datasets: SODA, MOCS, and CIS. To ensure comparability and fairness in the experiments, it applies the same hyperparameter configuration to each dataset. Specifically, it uses the configuration shown in Table \ref{tab:1}.
\begin{table}[]
\renewcommand\arraystretch{1.25}
\caption{Detailed experimental configuration.\label{tab:1}}
\begin{tabular}{cc}
\hline
Base learning rate & 0.01                             \\
Base weight decay  & 0.0005                           \\
Optimizer momentum & 0.937                            \\
Batch size         & 16                               \\
Training epochs    & 100                              \\
Warmup iterations  & max(1000, 3 * iters\_per\_epochs) \\
Input size         & 640*640                          \\ \hline
\end{tabular}
\end{table}

With the above strict configuration and selection criteria, these experiments can obtain reliable and trustworthy results, enabling the comparison of performance across different models and providing a solid foundation and reference for further research.

\subsection{Comparison with the SOTA Results}

\textit{1) Results on various image detection benchmarks: }This study conducts experiments on the SODA, MOCS, and CIS datasets to evaluate the performance of GSO-YOLO. The experimental results are summarized in Table \ref{tab:2}. From the results, it is evident that the proposed GSO-YOLO exhibits significant improvements in performance under various methodologies.

On the SODA dataset, the experimental results indicate that the proposed method, GSO-YOLO, demonstrates notable advantages over the baseline method when using YOLOv8 as the comparison baseline. At an IoU of 0.5, the mAP of GSO-YOLO increases from 70.13\% to 81.54\% compared to the baseline method. Furthermore, within the IoU range of 0.5 to 0.95, the mAP increases from 40.49\% to 50.20\%, signifying significant performance enhancements of 16\% and 24\%, respectively.

As the experiments extend to include other datasets, GSO-YOLO consistently outperforms the baseline method. Specifically, on the MOCS dataset, GSO-YOLO achieves significant improvements of 17\% and 24\% in mAP50 and mAP50-95, respectively, compared to the baseline method. Similarly, on the CIS dataset, GSO-YOLO also demonstrates slight enhancements, with performance improvements of 6\% and 17\% across different evaluation metrics.

These experimental results validate the significant effectiveness of GSO-YOLO in improving the performance of deep-learning models. GSO-YOLO proves its capability in substantially bridging existing performance gaps, highlighting its potential in enhancing the training process and optimizing the performance of deep learning models across various datasets.
\begin{table*}[!h]
\renewcommand\arraystretch{1.25}
\setlength{\tabcolsep}{16pt}
\caption{Comparison of YOLOv8 and GSO-YOLO on the construction site target detection datasets.\label{tab:2}}
\centering
\begin{tabular}{cccc}
\hline
Dataset               & Method   & mAP50             & mAP50-95          \\ \hline
\multirow{2}{*}{SODA} & YOLOv8   & 70.13\%           & 40.49\%           \\  
                      & \textbf{GSO-YOLO} & \textbf{81.54\%(↑11.41\%)} & \textbf{50.20\%(↑9.71\%)}  \\ \hline
\multirow{2}{*}{MOCS} & YOLOv8   & 64.02\%           & 46.24\%           \\ 
                      & \textbf{GSO-YOLO} & \textbf{75.13\%(↑11.11\%)} & \textbf{57.47\%(↑11.23\%)} \\ \hline
\multirow{2}{*}{CIS}  & YOLOv8   & 82.57\%           & 63.56\%           \\  
                      & \textbf{GSO-YOLO} & \textbf{88.03\%(↑5.46\%) } & \textbf{74.20\%(↑10.64\%)} \\ \hline
\end{tabular}
\end{table*}

\begin{figure}[!t]
\centering
\includegraphics[width=3.43in]{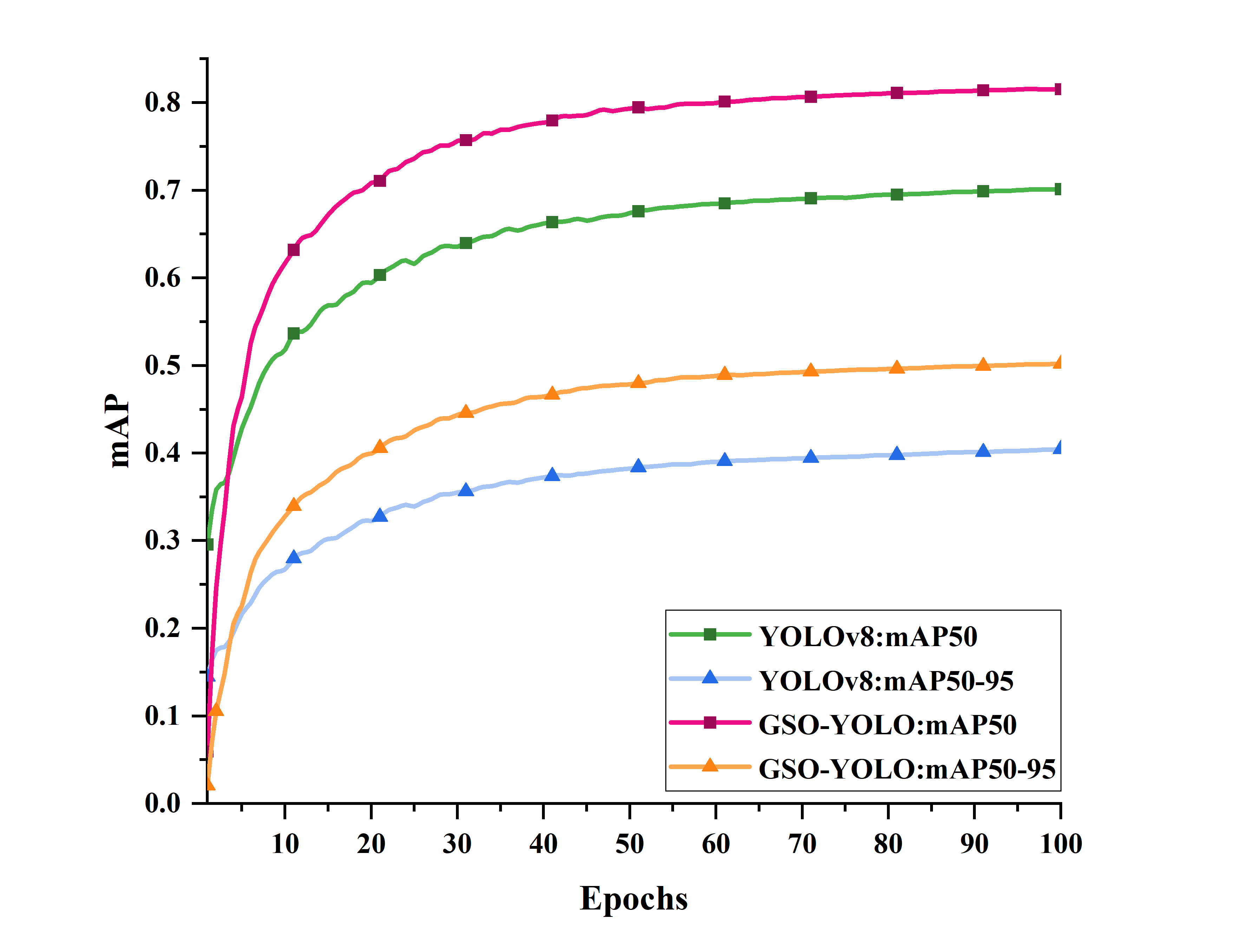}%
\caption{mAP-Epochs curves.}
\label{fig:3}
\end{figure}

\textit{2) Training-Accuracy Curve Analysis: }To illustrate the impact of the proposed method on training performance, this study conducts a comparative analysis between the baseline method and GSO-YOLO by using SODA. It visualizes the mAP-Epochs curves with mAP50, and the results are shown in Figure \ref{fig:3}. GSO-YOLO consistently outperforms the baseline method.

\begin{figure*}[!h]
\centering
\subfloat[]{\includegraphics[width=2.13in]{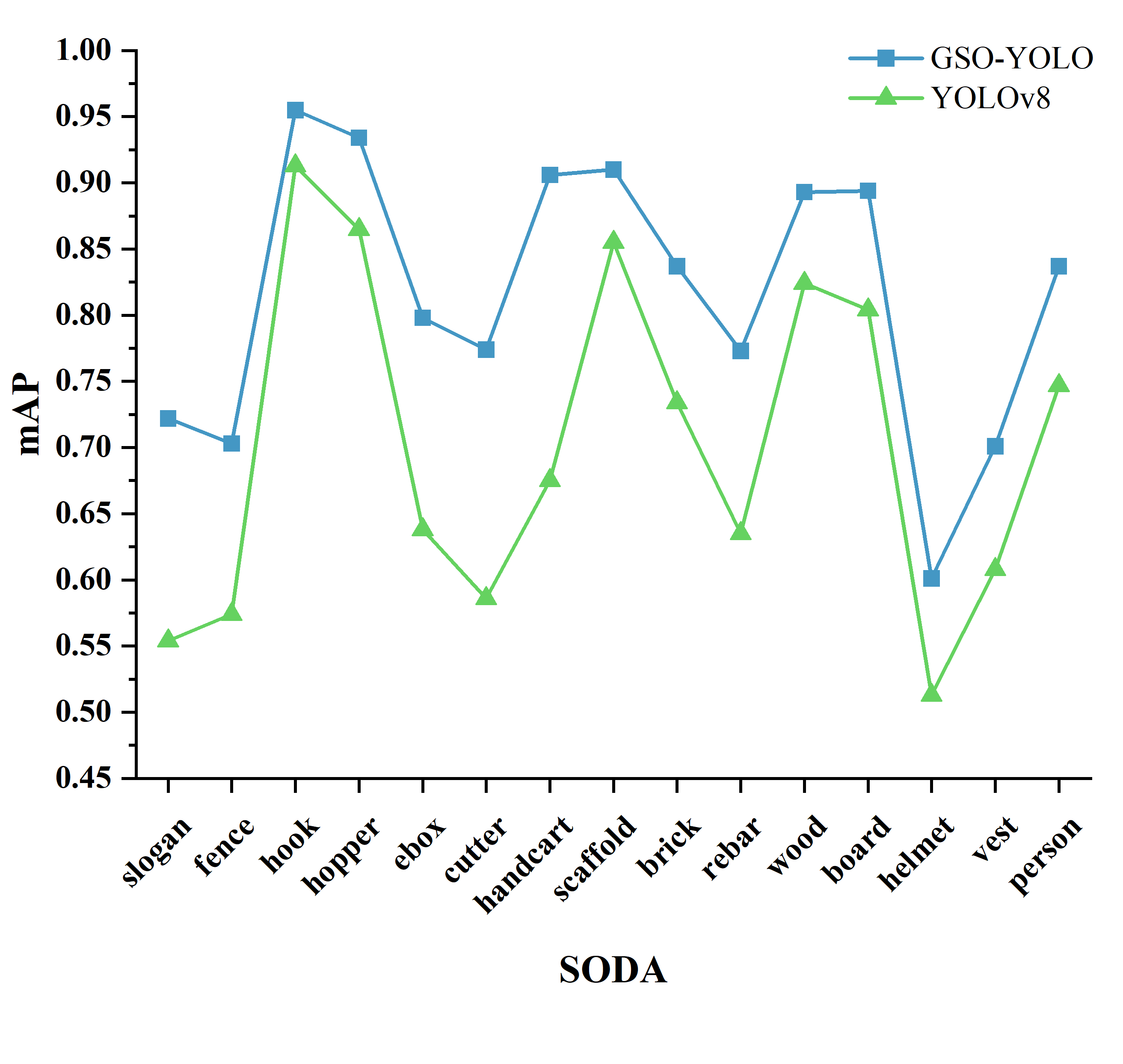}}%
\hfil
\subfloat[]{\includegraphics[width=2.13in]{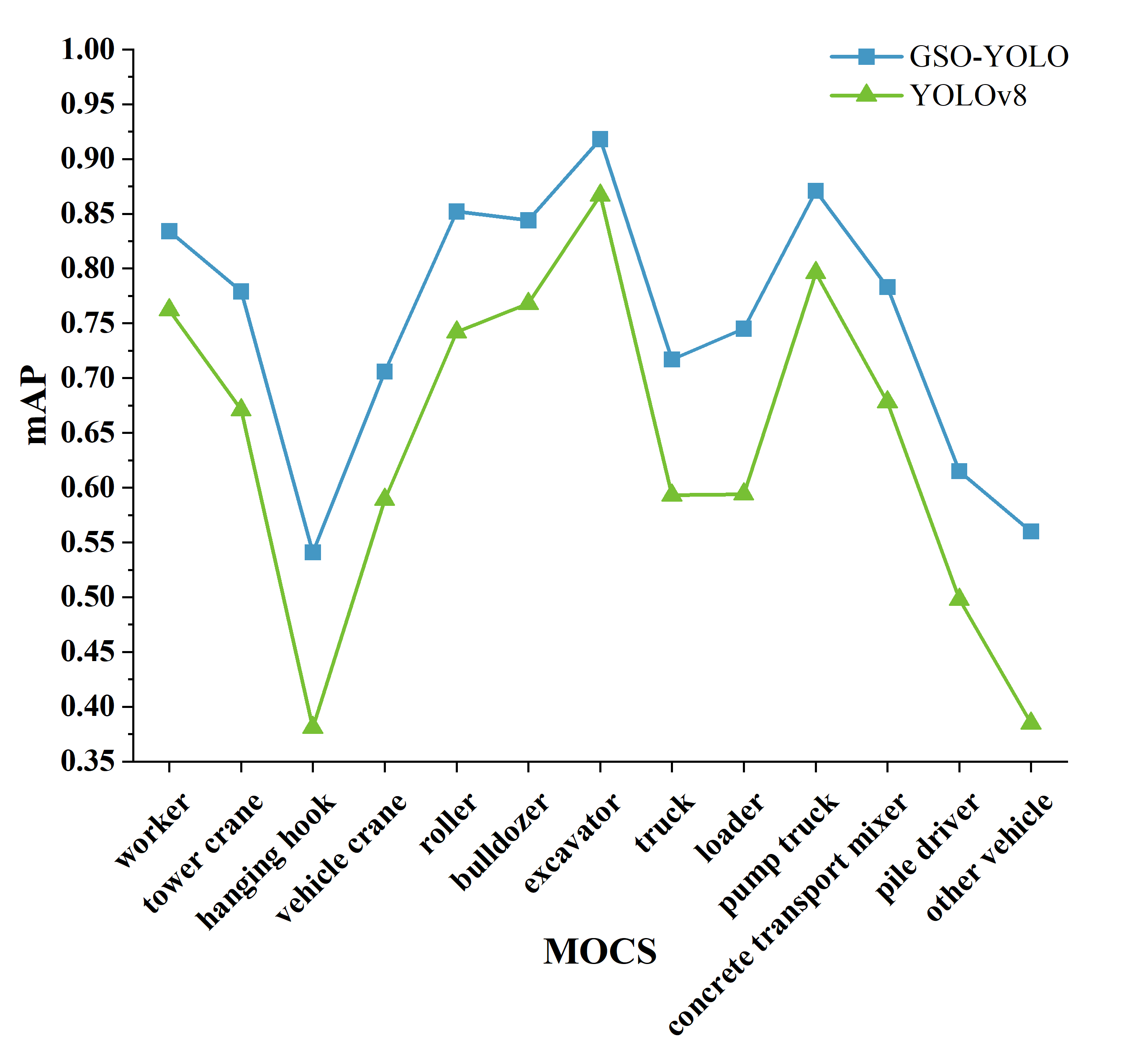}}%
\hfil
\subfloat[]{\includegraphics[width=2.13in]{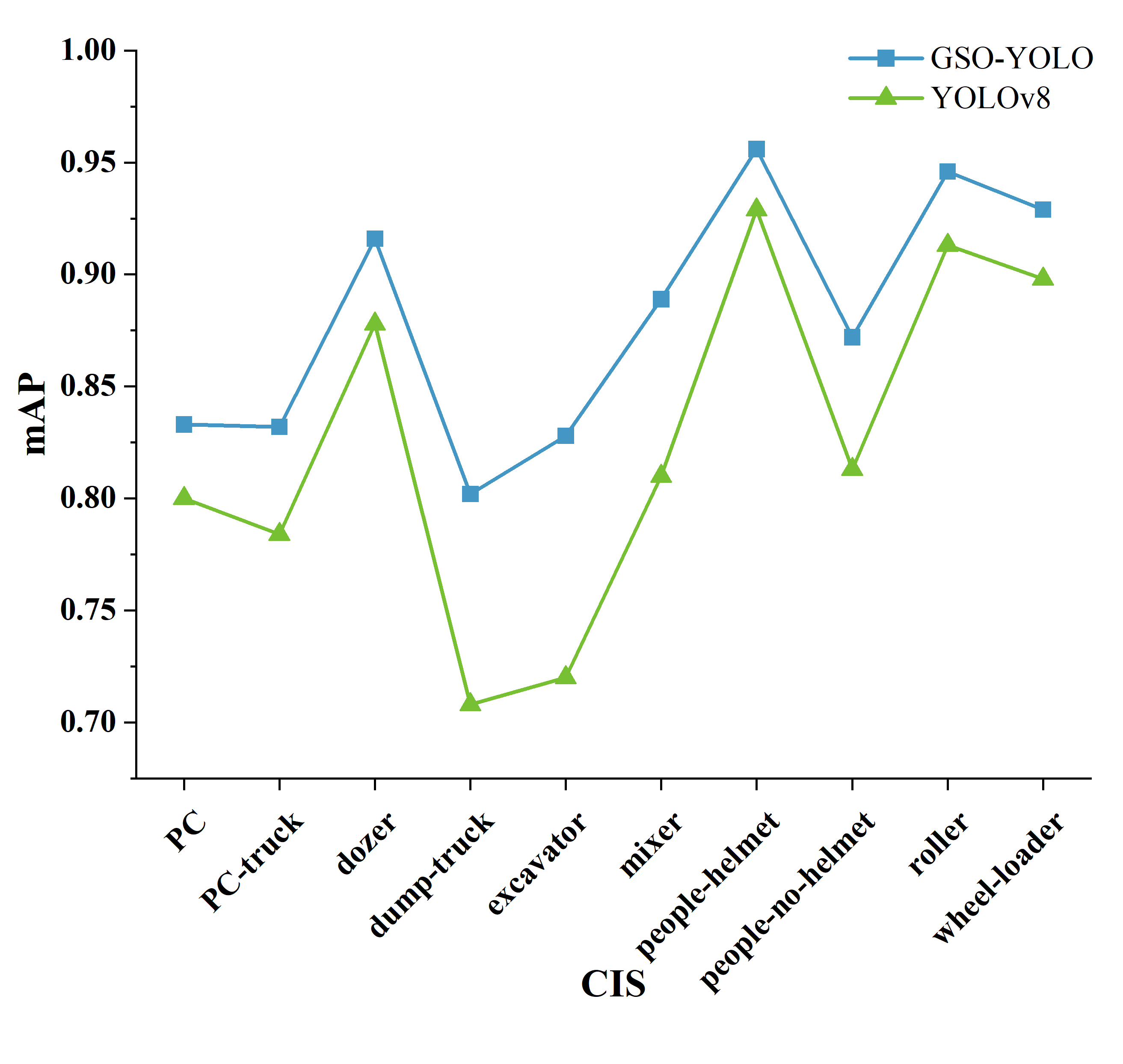}}%
\caption{mAP-classes curves. (a)mAP-classes curves on SODA. (b) mAP-classes curves on MOCS. (c) mAP-classes curves on CIS.}
\label{fig:4}
\end{figure*}

\textit{3) Classes-Accuracy Curve Analysis: }To compare the detection performance of different datasets within the same class, this study conducts a correlation analysis with mAP50. As illustrated in Figure \ref{fig:4}, it is evident that each class in every dataset performs better on GSO-YOLO compared to YOLOv8.
\subsection{Ablation Study}
\textit{1) Performance Comparison between YOLOv8 with GOM, SCM, and GSO-YOLO: }To further elucidate the contributions of GOM and SCM to GSO-YOLO, this study conducts experiments using YOLOv8 as the baseline method and sequentially adds GOM, SCM, and GSO-YOLO to the baseline method on the three datasets. The experimental results are summarized in Table \ref{tab:3}.
\begin{table*}[!h]
\renewcommand\arraystretch{1.25}
\setlength{\tabcolsep}{16pt}
\caption{Results of ablation experiments using different methods based on YOLOv8.\label{tab:3}}
\centering
\begin{tabular}{cccc}
\hline
Dataset               & Method         & mAP50             & mAP50-95          \\ \hline
\multirow{5}{*}{SODA} & YOLOv8         & 70.13\%           & 40.49\%           \\
                      & YOLOv8+SCM     & 72.06\%(↑1.93\%)  & 42.30\%(↑1.81\%)  \\
                      & YOLOv8+GOM     & 78.31\%(↑8.18\%)  & 46.77\%(↑6.28\%)  \\
                      & YOLOv8+SCM+GOM & 78.78\%(↑8.65\%)  & 47.12\%(↑6.63\%)  \\
                      & \textbf{GSO-YOLO}       & \textbf{81.54\%(↑11.41\%)} & \textbf{50.20\%(↑9.71\%)}  \\ \hline
\multirow{5}{*}{MOCS} & YOLOv8         & 64.02\%           & 46.24\%           \\
                      & YOLOv8+SCM     & 64.60\%(↑0.58\%)  & 46.61\%(↑0.37\%)  \\
                      & YOLOv8+GOM     & 71.86\%(↑7.84\%)  & 53.53\%(↑7.29\%)  \\
                      & YOLOv8+SCM+GOM & 72.06\%(↑8.04\%)  & 53.77\%(↑7.53\%)  \\
                      & \textbf{GSO-YOLO}       & \textbf{75.13\%(↑11.11\%)} & \textbf{57.47\%(↑11.23\%)} \\ \hline
\multirow{5}{*}{CIS}  & YOLOv8         & 82.57\%           & 63.56\%           \\
                      & YOLOv8+SCM     & 82.80\%(↑0.23\%)  & 63.57\%(↑0.01\%)  \\
                      & YOLOv8+GOM     & 87.47\%(↑4.90\%)  & 70.96\%(↑7.40\%)  \\
                      & YOLOv8+SCM+GOM & 87.51\%(↑4.94\%)  & 71.03\%(↑7.47\%)  \\
                      & \textbf{GSO-YOLO}       & \textbf{88.03\%(↑5.46\%)}  & \textbf{74.20\%(↑10.64\%)} \\ \hline
\end{tabular}
\end{table*}

From the results in the table, it is evident that, for example, on the SODA dataset, without GOM and SCM, the mAP50 and mAP50-95 only reach 70.13\% and 40.49\%, respectively. When SCM is added alone, the accuracy increases to 72.06\% and 42.30\%. Similarly, with the addition of GOM alone, the accuracy shows a significant improvement, reaching 78.31\% and 46.77\%. When both modules are combined, the mAP further increases to 78.78\% and 47.12\%. Notably, when both modules work together and the original loss function is replaced with AIoU, the accuracy further improves to 81.54\% and 50.20\%.

These findings provide compelling evidence of the significant contributions made by GOM, SCM, and AIoU, highlighting the remarkable performance achieved by GSO-YOLO through their integration.

\begin{figure}[!h]
\centering
\includegraphics[width=2.5in]{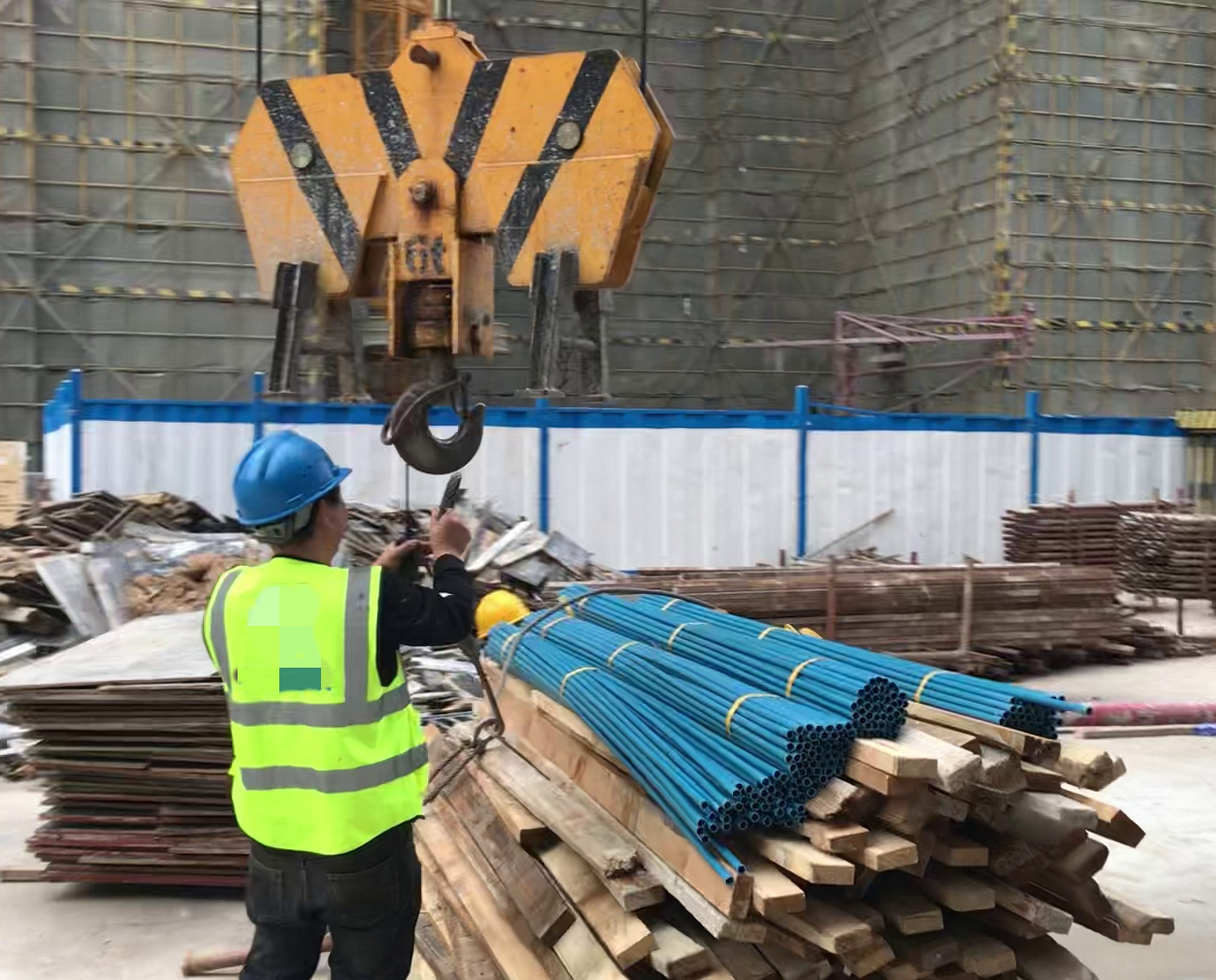}%
\caption{Original image example.}
\label{fig:5}
\end{figure}

\begin{figure*}[!h]
\centering
\subfloat[]{\includegraphics[width=1.23in]{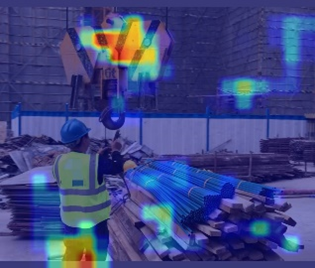}}%
\hfil
\subfloat[]{\includegraphics[width=1.23in]{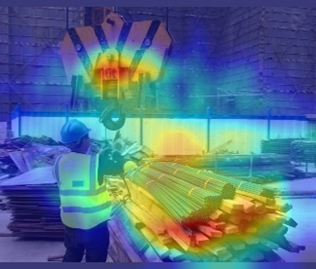}}%
\hfil
\subfloat[]{\includegraphics[width=1.23in]{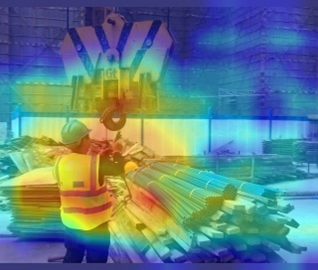}}%
\hfil 
\subfloat[]{\includegraphics[width=1.23in]{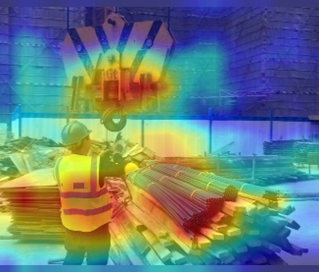}}%
\hfil
\subfloat[]{\includegraphics[width=1.23in]{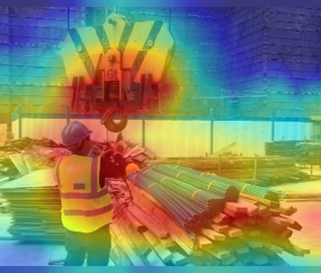}}%
\caption{Visualization of channel feature maps. (a) Feature Map of YOLOv8. (b) Feature Map of YOLOv8 with SCM. (c) Feature Map of YOLOv8 with GOM. (d) Feature Map of YOLOv8 with SCM and GOM. (e) Feature Map of GSO-YOLO.}
\label{fig:6}
\end{figure*}

\textit{2) Comparison of Features in Different Methods: }The research aims to demonstrate the significant contributions of the proposed GSO-YOLO to the field of construction detection and its competitive performance compared to YOLOv8. To showcase the advanced features of GSO-YOLO, this study conducts feature map analysis for Figure\ref{fig:5} on different configurations, including YOLOv8 with SCM, YOLOv8 with GOM, both modules applied simultaneously, and GSO-YOLO. In the heatmaps, deeper red indicates higher attention concentration. A larger red area represents a broader receptive field of the model. The detailed results of these feature maps are shown in Figure \ref{fig:6}. From left to right, the five feature maps represent five methods: YOLOv8, YOLOv8+SCM, YOLOv8+GOM, YOLOv8+SCM+GOM, and GSO-YOLO.

\begin{figure*}[!h]
\centering
\subfloat[]{\includegraphics[width=6.51in]{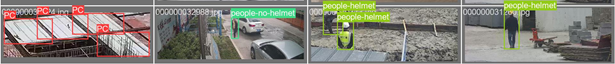}}%
\hfil
\subfloat[]{\includegraphics[width=6.51in]{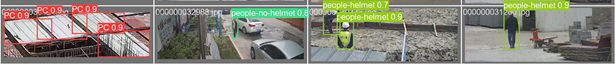}}%
\hfil
\subfloat[]{\includegraphics[width=6.51in]{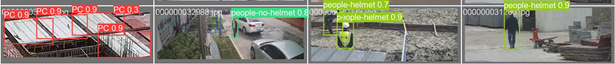}}%
\hfil
\subfloat[]{\includegraphics[width=6.51in]{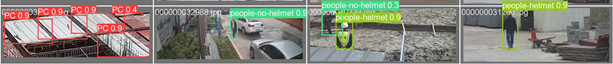}}%
\hfil
\subfloat[]{\includegraphics[width=6.51in]{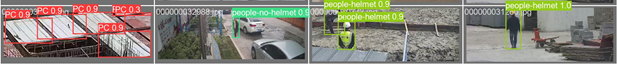}}%
\hfil
\subfloat[]{\includegraphics[width=6.51in]{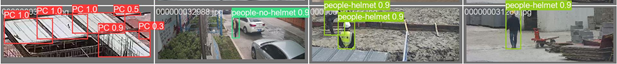}}%
\caption{Generalization Experimental Performance. (a) Ground truth labels. (b) Predicted labels using YOLOv8. (c) Predicted labels using YOLOv8+SCM. (d) Predicted labels using YOLOv8+GOM. (e) Predicted labels using YOLOv8+SCM+GOM. (f) Predicted labels using GSO-YOLO.}
\label{fig:7}
\end{figure*}

When analyzing the feature maps, (a) concentrates on a very limited area, indicating that YOLOv8 has limited effectiveness in object detection for construction scenes. In (b), with the addition of SCM, more object features are successfully captured. Similarly, with the involvement of GOM in (c), more global semantic features are noted. Notably, in (d), with both modules combined, GSO-YOLO captures not only more features of the objects to be identified but also shows significant performance enhancement after global optimization. Finally, with the application of AIoU in (e), the model improves both the sensitivity to detecting small objects and robustness, allowing it to better adapt to complex environmental changes. 

Therefore, GSO-YOLO systematically and strategically addresses the shortcomings of the original methods step by step.

\subsection{Generalization Study}
This section aims to investigate the generalization performance of the proposed GSO-YOLO and assess its effectiveness. This study trains the model using three datasets and tests it on their respective validation sets. The results, as shown in Figure \ref{fig:7}, demonstrate the impact of adding different modules on generalization ability compared to the ground truth labels in (a). By introducing GSO-YOLO, a significant expansion of the model's receptive field and a notable improvement in test accuracy can be observed, with the test precision for certain categories even reaching 1.0. These findings highlight the effectiveness of GSO-YOLO in enhancing the generalization capability of supervised learning, thereby improving its overall performance in construction scene object detection tasks.

\section{Conclusion}
This study develops the GSO-YOLO model for construction site monitoring integrated with GOM and SCM. GOM combines global attention weighting with local information to prevent information loss and enhance global interaction feature capture, improving the model's performance and generalization. SCM uses dynamic exponential moving averages to process historical detection results, reducing noise, smoothing outputs, and boosting stability and robustness. Together, GOM and SCM comprehensively enhance network performance.

Additionally, the AIoU enhanced loss function, merging CIoU and EIoU, improves detection in complex construction environments. CIoU focuses on the completeness of intersection and union between detection boxes, while EIoU emphasizes computational efficiency. Thus, AIoU ensures accuracy and efficiency, allowing GSO-YOLO to maintain high accuracy and efficient resource use.

These improvements are validated on the SODA, MOCS, and CIS construction site datasets, achieving state-of-the-art (SOTA) performance and showing that GSO-YOLO significantly enhances target detection performance.  In summary, GSO-YOLO effectively addresses complex construction site challenges, improving detection accuracy and stability.\\






\newpage









\bibliographystyle{unsrt}
\bibliography{main.bib}



\end{document}